\documentclass[runningheads]{llncs}
\usepackage[T1]{fontenc}

\usepackage{graphicx,verbatim}
\usepackage{tikz,amsmath,amssymb,bm}
\usetikzlibrary{arrows.meta,positioning,calc,decorations.pathreplacing}
\usepackage{booktabs,multirow,xcolor}
\usetikzlibrary{arrows.meta,positioning,calc}

\usepackage{booktabs}
\usepackage{booktabs,multirow,xcolor}
\usepackage{colortbl}
\usepackage{hyperref}
\usepackage{multirow}
\usepackage{rotating}
\usepackage{orcidlink}
\usepackage{colortbl} 

\begin{document}

\title{Self-Supervised Cardiac Phase Detection via Single-Parameter Latent Orbits}
\titlerunning{Cardiac Phase Detection via Latent Orbits}

\author{John Bonnici\inst{1}\orcidlink{0009-0002-3725-5034} \and
Matthew Baugh\inst{1}\orcidlink{0000-0001-6252-7658} \and
Aleksandra Kulbaka\inst{1}\orcidlink{0000-0002-4435-8827} \and \\
Sarah Cechnicka\inst{1}\orcidlink{0009-0008-3449-9379} \and
Bernhard Kainz\inst{1, 2}\orcidlink{0000-0002-7813-5023} \and
Alberto Gomez\inst{3}\orcidlink{0000-0002-7897-7589}}
\authorrunning{J. Bonnici et al.}

\institute{Departament of Computing, Imperial College London, UK
\email{john.bonnici21@imperial.ac.uk} \and
Friedrich–Alexander University Erlangen–N\"urnberg, DE \and
Ultromics Ltd, Oxford, UK\\
}

\maketitle              
\begin{abstract}
Accurate identification of end-diastole (ED) and end-systole (ES) in echocardiography underpins the quantification of ventricular function, yet manual selection of these key frames is subjective and introduces clinically significant inter-operator variability.
Recent self-supervised methods either prescribe strict periodic trajectories or learn an unconstrained low-dimensional motion subspace from reconstruction or registration objectives. The former offers interpretability but imposes restrictive assumptions on temporal progression, whereas the latter leaves cardiac phase implicit and ED/ES must be recovered through post-hoc geometric processing of the learned trajectory.
We translate the physiological observation that cardiac phase is a one-dimensional signal into a prior by constraining the latent motion component to a \emph{single-parameter latent orbit}, \emph{i.e.}, a global linear trajectory in latent space indexed by a bounded scalar phase variable. 
Mapping this variable through a sinusoidal nonlinearity yields an oscillatory motion signal with consistent temporal ordering, enabling direct identification of ED and ES from the learned phase signal. This inductive bias allows the model to capture an interpretable representation of the cardiac cycle, while maintaining flexibility to capture irregular heartbeats. 
Trained on EchoNet-Dynamic without annotations, our minimal single-parameter cardiac phase model learns an effective latent orbit, significantly improves upon the previous state of the art in ED localisation and matches it in ES localisation while using a more constrained representation and fewer training epochs. This demonstrates that a principled physiological inductive bias can match or exceed the performance of more complex representations. Code is available at: \href{https://github.com/BonniciJ/OrbitalEcho/}{https://github.com/BonniciJ/OrbitalEcho/}


\end{abstract}

\section{Introduction}
Cardiovascular disease is the leading cause of death worldwide~\cite{martin2025aha,who2025cvd}. Transthoracic echocardiography is the primary imaging modality for cardiac assessment due to its portability, low cost, and real-time acquisition~\cite{lang2015recommendations}. Cardiac function is often quantified using left ventricular ejection fraction (LVEF), which measures the proportion of blood ejected from the left ventricle during each heartbeat. Estimating LVEF requires accurate identification of end-diastole (ED) and end-systole (ES) frames~\cite{lang2015recommendations,mada2015define}, yet manual selection is subjective, with substantial inter-operator ejection-fraction variability~\cite{otterstad1997accuracy,thavendiranathan2013reproducibility}. 
Automated, annotation-free ED/ES detection could therefore improve clinical reproducibility and the scalability of large cohort studies.

Recent work on self-supervised motion-structure decomposition has shown that an autoencoder trained purely on video reconstruction can separate a static structural component from a low-dimensional motion trajectory, and that cardiac phase can be recovered from the geometry of that trajectory without any labels~\cite{yangLatentMotionProfiling2026,yangOrientationRobustLatentMotion2026}. However, existing formulations leave the motion coefficients unconstrained, allowing points to arbitrarily span the low-rank subspace~\cite{rohe2018low,mcleod2015spatiotemporal}. As a result, phase extraction requires post-hoc signal processing, such as principal-axis projection or RANSAC-based orientation detection.

We address the same problem using a strong physiological prior implemented directly in the disentangled latent space. In the apical four-chamber view (A4C), cardiac motion presents multiple simultaneous patterns: LV, RV, and atrial contraction and relaxation, valve motion, and out-of-plane motion of some structures; however, cardiac phase is a scalar signal that determines the temporal position within the cycle. 
In contrast with competing methods, we replace a multi-dimensional latent motion representation with a \emph{single-parameter latent orbit} - a line segment in latent space indexed by a bounded scalar phase variable - and propose a latent disentanglement approach that promotes phase separation.
Our contributions are as follows:

\textbf{(1)} 
    We propose a single-parameter latent orbit model that encodes cardiac phase as a cyclic traversal of a one-dimensional trajectory.
    \textbf{(2)} 
    We show that ED and ES can be retrieved directly from the minima and maxima of the learned phase signal.
    \textbf{(3)} 
    Trained on EchoNet-Dynamic without ED or ES labels, we show that our model improves ED localisation and matches ES localisation while using a more constrained representation and fewer training epochs, achieving strong downstream EF prediction performance and cycle consistency with a principled, minimal-parameter design. 

\noindent\textbf{Related work.}
Supervised ED/ES detection methods use CNN-RNN hybrids ~\cite{lee2020automatic,li2023semi,dezaki2019cardiac}, vision transformers~\cite{reynaud2021uvt}, multi-task fetal networks~\cite{pu2024hfsccd}, and indirect derivation from volume curves or segmentation masks~\cite{ouyangVideobasedAIBeattobeat2020a,zeng2023maef}, but require costly annotations. Among unsupervised approaches, early manifold-learning methods used locally linear embedding to detect phase from trajectory density~\cite{gifani2010automatic,shalbaf2015echo}, while the training-free DDSB method~\cite{bu2025ddsb} exploited expansion--contraction dynamics but was sensitive to short ED/ES intervals. DeepHeartBeat~\cite{laumerDeepHeartBeatLatentTrajectory} constrained motion to a circular trajectory with fixed frequency and shape, improving periodicity but limiting flexibility to beat-to-beat variability. Latent motion profiling (LMP)~\cite{yangLatentMotionProfiling2026} introduced self-supervised motion-structure decomposition via a joint task of frame-wise and mean frame reconstruction, using an unconstrained two-dimensional motion subspace. LMP achieved state-of-the-art performance among annotation-free ED/ES detection methods when tested on EchoNet-Dynamic. Its orientation-robust extension ORBIT~\cite{yangOrientationRobustLatentMotion2026} adapted this paradigm to fetal imaging using registration-based self-supervision.

We build on the motion-structure decomposition of~\cite{yangLatentMotionProfiling2026} but replace unconstrained motion coefficients with a single-parameter orbit derived from the prior that cardiac phase is one-dimensional.

\section{Method}

\begin{figure}[t]
\centering
\resizebox{\textwidth}{!}{
\includegraphics{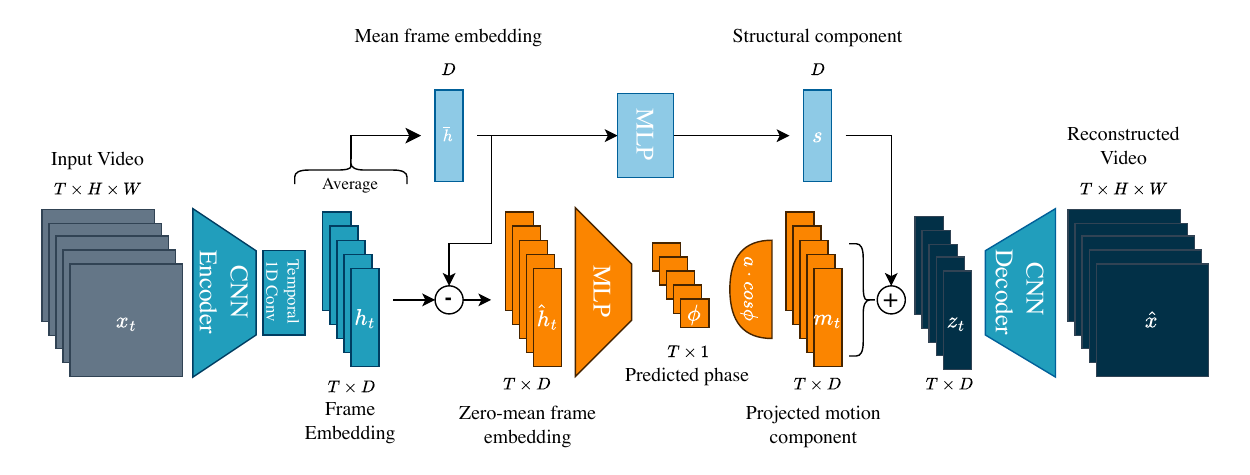}
}
\caption{
Each frame in the input clip is encoded into a latent space and temporally mixed. This representation is processed by two branches:
the \emph{structure branch} (top) computes the clip-level mean and maps it to a static latent $\bm{s}$; and 
the \emph{motion branch} (bottom) predicts frame-level phase values $\phi_t$, projecting through a learned sinusoidal vector $\bm{a}\in\mathbb{R}^{D}$ into the motion latent $\bm{m}_t$.
The latent $z_t = \bm{s} + \bm{m}_t$ is decoded to
reconstruct the input.}
\label{fig:overview}
\end{figure}

Given a training set of A4C echocardiography videos, we sample a $H \times W$ clip of $T$ consecutive greyscale frames, $\mathbf{x}=\{x_t\}_{t=1}^{T}$, $x_t \in \mathbb{R}^{H\times W}$ from each video (with a random starting frame). These clips are used to train an autoencoder whose latent space is explicitly decomposed into a time-invariant \emph{structure} component and a time-varying \emph{motion} component constrained to a one-dimensional parametrised orbit. The overall architecture is shown in Fig.~\ref{fig:overview}.

Each frame is independently encoded by a standard convolutional backbone, $\tilde{h}_t = f_{\text{enc}}(x_t) \in \mathbb{R}^{C}$. To incorporate local temporal context without a heavy sequence model, we apply a lightweight 1-D convolution over time, with kernel size~$5$, yielding temporally mixed features $h_{1:T} = f_{\text{mix}}(\tilde{h}_{1:T})$, $h_t \in \mathbb{R}^{C}$. 

Assuming that cardiac phase is captured in the encoded frames as a zero-mean oscillation, temporal averaging of the encoded frames effectively cancels the dominant cardiac motion information. We therefore compute the clip-level mean $\bar{h} = \frac{1}{T}\sum_{t=1}^{T} h_t$ and project it through a learned multilayer perceptron (MLP) to obtain the structure latent $\bm{s} = f_{\text{str}}(\bar{h}) \in \mathbb{R}^{D}$. This vector captures patient-specific anatomy and is shared across all frames in the clip.
By subtracting the latent clip mean from the mixed frames, $\hat{h}_t = h_t - \bar{h}$, we remove the stationary component, producing a latent that contains dynamic information (motion, cardiac phase) as well as high-frequency features (e.g. speckle, edges) that are not captured by $\bar{h}$.

The key design choice is how to map $\hat{h}_t$ to the scalar phase signal where ED and ES can be detected as salient events. Prior work~\cite{yangLatentMotionProfiling2026} used a 2D subspace where motion, phase, and other dynamic features are entangled, with the phase signal extracted via PCA. We instead project $\hat{h}_t$ into a 1D signal directly via an MLP, under the constraint that this signal is cyclic using a sinusoidal projection to bring it back to a $D$-dimensional space.
This way, we predict a bounded scalar phase variable $\phi_t$ that indexes the motion state. We obtain an oscillatory motion signal by applying a sinusoidal nonlinearity, while retaining flexibility to accommodate beat-to-beat variability in instantaneous heart rate. Phase is predicted as $\phi_t = \pi \cdot \tanh\!\bigl(f_{\text{phase}}(\hat{h}_t)\bigr)$, where $f_{\text{phase}}$ is a small MLP.

Given the scalar phase variable $\phi_t$, we construct the motion latent by mapping
through a cosine basis and a single learned, global direction
$\bm{a}\in\mathbb{R}^{D}$: 
$
  \bm{m}_t = \bm{a} \,\cos(\phi_t)  \in \mathbb{R}^{D}.
$
The motion component therefore oscillates along a single axis in the $D$-dimensional latent space. Geometrically, as $\phi_t$ advances through a cycle, $\bm{m}_t$ oscillates along a line segment according to the predicted phase signal. 
The two extrema correspond to maximal relaxation and contraction of the LV, \emph{i.e.}, end-diastole and end-systole: intermediate frames look similar whether contracting or relaxing and map to the same orbit position, while ED and ES are maximally distinct and occupy the extrema.

Because the learned direction $\bm{a}$ is unconstrained in $\mathbb{R}^D$, the scalar phase drives coordinated changes across anatomical features (wall thickness, valve position, chamber area) while the decoder retains full expressivity; sharing $\bm{a}$ across samples induces cycle consistency, placing ED and ES on distinct regions of the line across all videos.

The per-frame latent is formed by simple addition of the structure and motion components, $z_t = \bm{s} + \bm{m}_t$. This per-frame latent is decoded by a convolutional decoder $\hat{x}_t = g_\theta(z_t)$ to estimate the input frame.

\noindent\textbf{Cardiac phase identification.}\label{sec:edid}
ED and ES correspond to the largest and the smallest states of the left ventricle, therefore these two events are, in the latent space, as far as possible from the average state (captured by  $\bar{h}$). As a result, we hypothesise that ED and ES correspond to the extrema of the phase values, in a cycle-consistent manner across videos, with the assignment fixed by the sign convention of $\bm{a}$. We determine this assignment once on the validation set and apply it at test time. 
We apply a high-pass filter to remove low-frequency drift, and a smoothing filter on the normalised phase signal. We exclude events that do not follow an alternating ED/ES pattern.

\noindent\textbf{Training objective.}
We train the model end-to-end using only the pixel-space similarity loss between frame-wise reconstructions and input images. We apply a Gaussian blur to the input images before computing the similarity loss to isolate phase information from other sources of inter-frame variation, such as high-frequency speckle and fine edge definition, which vary independently of cardiac phase. 
$\mathcal{L}
  = \frac{1}{T}\sum_{t=1}^{T}
    \lVert \hat{x}_t - \mathcal{G}_\sigma(x_t)\rVert_2^2$,
where $\mathcal{G}_\sigma$ is a fixed Gaussian blur operator.

\section{Experiments and Results}\label{sec:experiments}

We evaluate our model on the EchoNet-Dynamic dataset~\cite{ouyangVideobasedAIBeattobeat2020a}, which contains 10,030 A4C echocardiography videos with expert-annotated ED and ES frames. We adopt the official train/validation/test split and withhold all annotations during training. We train using three NVIDIA RTX A6000 GPUs with all numerical results reported after only 50 epochs of training. Following prior work~\cite{yangLatentMotionProfiling2026}, we report the mean absolute error (MAE) in both frames and milliseconds between the predicted and the closest ground-truth ED/ES label. Additionally, we also train an EF regressor on the learned representations to evaluate their practical utility. All metrics are computed over full-length videos at native frame rate. We compare against supervised~\cite{zeng2023maef,li2023semi} and unsupervised baselines~\cite{yangLatentMotionProfiling2026,bu2025ddsb}.  For methods without published results on EchoNet-Dynamic, we re-run the authors' code with default hyperparameters on the same split; where code is unavailable we report numbers from the original publications and mark them accordingly.

\begin{table}[t]
\centering
\caption{%
ED/ES detection on EchoNet-Dynamic (test set).
MAE is reported in frames (fr.) and milliseconds (ms).
St. dev. on MAE reported in brackets.
EF results from MLP head trained for 100 epochs on learned features.
Best unsupervised result per column in \textbf{bold}.
$^{\dagger}$Reimplemented; $^{\ddagger}$from original publication.
DDSB results taken from \cite{bu2025ddsb,yangLatentMotionProfiling2026}.}
\label{tab:main}
\setlength{\tabcolsep}{3pt}
\renewcommand{\arraystretch}{1.1}
\resizebox{\textwidth}{!}{%
\begin{tabular}{@{} c l cc cc ccc @{}}
\toprule
& & \multicolumn{2}{c}{\textbf{ED MAE}} &
    \multicolumn{2}{c}{\textbf{ES MAE}} &
    \multicolumn{3}{c}{\textbf{EF}} \\
\cmidrule(lr){3-4}\cmidrule(lr){5-6}\cmidrule(lr){7-9}
& \textbf{Method} &
  fr. $\downarrow$ & ms $\downarrow$ &
  fr. $\downarrow$ & ms $\downarrow$ &
  MAE (\%) $\downarrow$ & RMSE (\%) $\downarrow$ & $R^2$ score $\uparrow$ \\
\midrule

\multirow{3}{*}{\rotatebox[origin=c]{90}{\tiny\begin{tabular}{@{}c@{}}\textit{Supervised}\\ \textit{(up. bound)}\end{tabular}}} 
 & Maani et al.~\cite{maani2024coreecho}$^{\ddagger}$
  & $--$ & $--$
  & $--$ & $--$
  & $3.90$ & $5.13$ & $0.82$ \\
 & Li et al.~\cite{li2023semi}$^{\ddagger}$
  & $2.1_{(2.5)}$ & $40.2$
  & $1.7_{(2.7)}$ & $32.6$
  & $--$ & $--$ & $--$ \\
 & MAEF~\cite{zeng2023maef}$^{\ddagger}$
  & $2.28_{(2.31)}$ & $44.5$
  & $2.43_{(2.16)}$ & $46.1$
  & $6.29$ & $8.21$ & $0.72$ \\
\midrule

\multirow{4}{*}{\rotatebox[origin=c]{90}{\tiny\begin{tabular}{@{}p{2.2cm}@{}}\centering \textit{Unsupervised /}\\ \textit{self-supervised}\end{tabular}}} 
 & DDSB~\cite{bu2025ddsb}$^{\ddagger}$
  & $4.3_{(4.8)}$ & $83.4$
  & $8.9_{(9.5)}$ & $175.4$
  & $--$ & $--$ & $--$ \\
 & LMP (500 epochs)~\cite{yangLatentMotionProfiling2026}$^{\dagger}$
  & $3.10_{(3.66)}$ & $60.1_{(69.0)}$
  & $2.21_{(3.45)}$ & $43.3_{(68.4)}$
  & $6.27$ & $8.24$ & $0.55$ \\[3pt]
 & LMP (50 epochs)~\cite{yangLatentMotionProfiling2026}$^{\dagger}$
  & $6.69_{(6.79)}$ & $131_{(131)}$
  & $6.46_{(7.60)}$ & $126_{(146)}$
  & $6.82$ & $8.89$ & $0.472$ \\[3pt]
 & \cellcolor{blue!5}\textbf{Ours (50 epochs)}
  & \cellcolor{blue!5}$\mathbf{2.36}_{(\mathbf{2.60})}$ & \cellcolor{blue!5}$\mathbf{{46.0_{(51.0)}}}$
  & \cellcolor{blue!5}$\mathbf{{2.13}_{(2.27)}}$ & \cellcolor{blue!5}$\mathbf{{{41.6}_{(44.5)}}}$
  & \cellcolor{blue!5}$\mathbf{6.02}$ & \cellcolor{blue!5}$\mathbf{8.07}$ & \cellcolor{blue!5}$\mathbf{0.56}$ \\
\midrule

 & \cellcolor{blue!5}\textbf{Ours (Bias Corrected)}
  & \cellcolor{blue!5}$\mathbf{2.17}_{(\mathbf{2.53})}$ & \cellcolor{blue!5}$\mathbf{42.5_{(49.5)}}$
  & \cellcolor{blue!5}$\mathbf{{2.12}_{(2.24)}}$ & \cellcolor{blue!5}$\mathbf{{41.4_{(43.9)}}}$
  & \cellcolor{blue!5}$--$ & \cellcolor{blue!5}$--$ & \cellcolor{blue!5}$--$ \\
\bottomrule
\end{tabular}}
\end{table}

Table~\ref{tab:main} summarises the ED/ES detection and EF prediction performance. Among unsupervised methods, our 1D orbit model achieves the lowest MAE, with an ED MAE of $2.36$ frames ($46.0$ ms) and an ES MAE of $2.13$ frames ($41.6$ ms). Compared with the previous SOTA (LMP~\cite{yangLatentMotionProfiling2026}), our approach significantly reduces ED MAE by $0.74$ frames on average, 95\% bootstrap CI [$0.57$, $0.92$], Wilcoxon signed-rank test with Bonferroni correction, $(p \ll 0.001)$. For ES, the mean improvement is smaller at $0.09$ frames, 95\% bootstrap CI [$-0.08$, $0.29$], so is not statistically significant. Our method did reduce the standard deviation on the error, indicating fewer outliers and more consistent predictions. The model also predicts a similar average number of ED and ES events per video to LMP~\cite{yangLatentMotionProfiling2026}, with 4.92 ED events per video (4.97 for LMP), and 4.26 ES events per video (4.05 for LMP), acting as an indirect proxy for false-positives. Together, these results support the use of the cyclical prior, particularly for ED detection. A direct comparison to DeepHeartBeat~\cite{laumerDeepHeartBeatLatentTrajectory} cannot be made because this method predicts ED/ES on the cycle itself, rather than in terms of frame error.

\noindent\textbf{Ejection Fraction.}
We train an MLP prediction head in a supervised manner (using the training set and EF labels from EchoNet-Dynamic) on the concatenated learned structure and motion features with the base model weights frozen.
We compare our approach with LMP~\cite{yangLatentMotionProfiling2026} by training an equivalent MLP head using the concatenation of its structure and motion features. Our method yields improved EF prediction performance ($6.02\%$ MAE vs.\ $6.27\%$ MAE) with one-tenth as many pre-training epochs, indicating that the proposed cyclic prior provides a strong inductive bias for capturing motion cues relevant to cardiac function assessment. Fully supervised methods (\emph{e.g.},  \cite{maani2024coreecho}) act as an upper bound.

\noindent\textbf{Bias Correction.}
We identify ED and ES by detecting extrema in our predicted phase signal, without any enforcement during training that these frames should lie at extrema. To assess this alignment, we used labelled data from 100 validation samples ($\sim7$\%) to compute the \emph{average signed error}. We observed mean biases of $-1.5$ frames for ED prediction and $-0.7$ frames for ES prediction. Correcting this systematic bias led to improved performance as shown in Table~\ref{tab:main}.

\begin{figure}[t]
    \centering
    \resizebox{\textwidth}{!}{
        \includegraphics{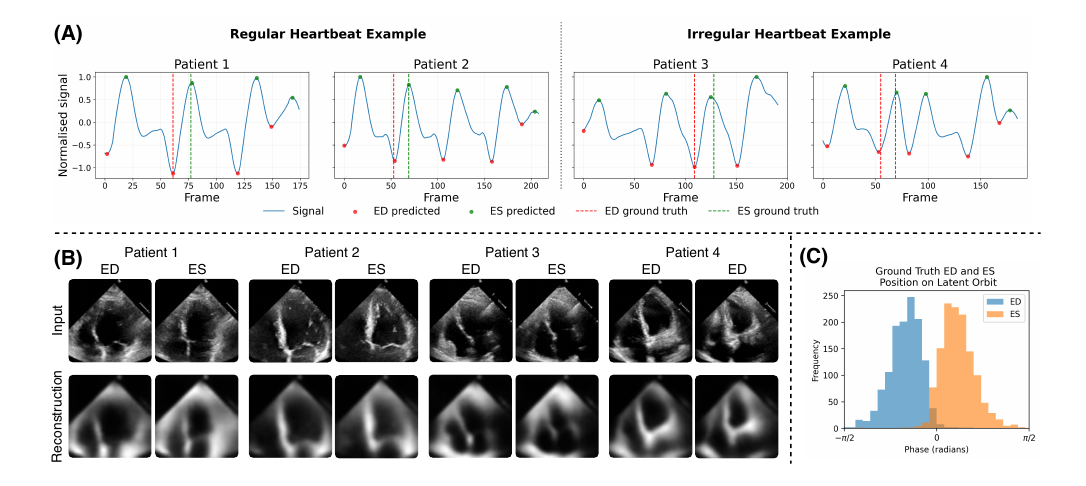}
    }
    \caption{%
        Qualitative results.
        \textbf{(A)} Predicted phase signal with predicted ED/ES frames vs ground truth for regular (left) and irregular (right) heartbeats.
        \textbf{(B)} Example reconstructions.
        \textbf{(C)} Histogram of where ground truth ED/ES lie on the orbit.
        }
    \label{fig:qualitative}
\end{figure}

\noindent\textbf{Qualitative analysis.}
Figure~\ref{fig:qualitative}~(A) shows the predicted phase signal for two pairs of patients: patients 1 \& 2 and 3 \& 4 exhibit regular and irregular heartbeats, respectively. The plots each show multiple ED and ES events, with ground truth labels available for a single heartbeat. Our model predicts a cyclic pattern with ED and ES located at the minima and maxima of the signal, respectively. The predicted ED/ES align closely with the expert-annotated ground truth. The irregular phase signal observed in patients 3 \& 4 shows variable gaps between ED and ES events across heartbeats. These samples were confirmed to be irregular by visual inspection of the dataset videos (please see  \href{https://bonnicij.github.io/OrbitalEcho/supplementary/}{supplementary material} for videos). 

Figure~\ref{fig:qualitative}~(B) shows the reconstructions for these four patients at ED and ES, generated purely from the learned latent orbit and the decoder. High-frequency speckle and noise are purposefully not captured by our representation with only low-frequency motion retained. These reconstructions are solely used for pre-training, and our objective is to capture cardiac phase, which is clearly represented in these reconstructions (please see  \href{https://bonnicij.github.io/OrbitalEcho/supplementary/}{supplementary material} for videos).  

Figure~\ref{fig:qualitative}~(C) shows the histogram of the positions of the ED/ES labels on the latent orbit, showing cycle consistency across patients (labels cluster to similar phase values) and separation of ED and ES regions of the orbit (little overlap of the two histograms). 

\begin{table}[t]
\centering
\caption{%
Ablation study.
MAE is reported in frames (fr.) and milliseconds (ms).
All models are trained for 50 epochs. Best are bolded, second-best are underlined.
} 
\label{tab:ablation}
\setlength{\tabcolsep}{4pt}
\small
\renewcommand{\arraystretch}{0.75}

\begin{tabular}{@{}l c c c c@{}}
\toprule
& \multicolumn{2}{c}{\textbf{ED MAE}} &
  \multicolumn{2}{c}{\textbf{ES MAE}} \\
\cmidrule(lr){2-3}\cmidrule(lr){4-5}
\textbf{Variant} &
  mean (fr.) $\downarrow$ & mean (ms) $\downarrow$ & 
  mean (fr.) $\downarrow$ & mean (ms) $\downarrow$ \\ 
\midrule
\rowcolor{blue!5}
\textbf{Full model (1D orbit)} &
  $\underline{{2.36}_{({2.60})}}$ & 
  $\underline{{46.0_{(51.0)}}}$ &  
  $\underline{{2.13}_{(2.27)}}$ & 
  $\underline{{41.6}_{(44.5)}}$  \\
w/o temporal mixing &
  $2.41_{(2.93)}$ &
  $47.1_{(57.9)}$ &
  $2.16_{(2.62)}$ & 
  $41.8_{(45.8)}$  \\
Two parameter orbit &
  $2.41_{(3.17)}$ &
  $47.1_{(63.0)}$ & 
  $2.18_{(2.69)}$ & 
  $42.4_{(47.9)}$\\
w/o Gaussian blur &
  $2.39_{(3.18)}$ &
  $46.8_{(63.4)}$ & 
  $\mathbf{2.11_{(2.54)}}$ & 
  $\mathbf{40.8_{(44.8)}}$ \\
Per-video parameter &
  $\mathbf{2.29_{(2.43)}}$ &
  $\mathbf{44.6_{(47.3)}}$ & 
  $2.55_{(3.04)}$ & 
  $49.5_{(52.7)}$  \\
\bottomrule
\end{tabular}
\end{table}

\noindent\textbf{Ablation study.}
Table~\ref{tab:ablation} isolates the contribution of each design choice. Removing the temporal mixing degrades performance on all metrics, as does extending to two parameters (discussed further in the next paragraph). Removing the Gaussian blur on the input frames reduces ED accuracy and slightly improves ES accuracy compared with the full model.
We also tested predicting the phase axis independently per video using an attention-weighted mean and the standard deviation of the motion latents as input to a small parameter prediction MLP. This resulted in a slightly improved ED but reduced ES performance, with reduced cycle consistency across videos; this parameter can point in any arbitrary direction for a given video, with any scale, removing the identifiability of ED and ES frames in the cycle. 
Overall we find the combination of blurring, temporal mixing, single-parameter orbit and a global parameter provide the best trade-off between ED and ES performance and stability.

\noindent\textbf{Extension to two parameters.} \label{par:2D extension}
The single-parameter orbit is our primary model. As an ablation we evaluate a two-parameter variant in which the cosine-sine pair $\bm{u}_t = [\cos\phi_t,\;\sin\phi_t]^{\!\top}$ is projected through a learned matrix $A\in\mathbb{R}^{2\times D}$ to yield an elliptical orbit $\bm{m}_t = \bm{u}_t^{\!\top}\! A$.
The elliptical formulation requires phase to be monotonic, such that the model predicts positive phase increments (via positive, sigmoid-gated increments) which are cumulatively summed to form the final phase signal. Furthermore, this formulation lacks cycle consistency across videos and ED/ES no longer cluster, hence making ED/ES disambiguation impossible. 
Without this monotonicity constraint, the model collapses to zero motion or to an oscillatory arc of the ellipse, a 1D approximation, yielding lower ED/ES accuracy as presented in Table~\ref{tab:ablation}.

\section{Discussion}\label{sec:discussion}
Our results support the hypothesis that A4C cardiac phase is intrinsically one-dimensional: a single parameter, combined with a learned linear projection, is sufficient to capture the contraction--relaxation dynamics and to localise ED and ES. We also show that using this learned orbital motion, we can train a lightweight EF regression MLP head. The ablation confirms that each component of the model design contributes meaningfully, and that the single-parameter representation and Gaussian blurring of the input both improve cycle consistency and ED/ES prediction accuracy; removing them hinders both temporal ordering and ED/ES identifiability. We note two practical considerations for future work. First, the current evaluation is limited to A4C videos from EchoNet-Dynamic; whether the one-dimensional assumption extends to parasternal or subcostal views remains to be validated. Second, while the model accommodates beat-to-beat heart-rate variability through per-frame phase prediction, highly irregular rhythms (\emph{e.g.}, atrial fibrillation with large cycle-length variance) warrant dedicated investigation. A limitation of the dataset used is that there is only one pair of ED and ES labels per video, and in consequence in some examples other erroneous predictions such as false positives are effectively ignored; this same limitation applies to baseline methods. 

\noindent\textbf{Bias Correction.}
Previous unsupervised methods, as well as the approach presented here, identify ED and ES frames as extrema of predicted or derived cardiac phase signals, or proxies thereof. However, these approaches generally do not justify why such extrema should align with the clinically defined conventions of ED and ES. Our results indicate that, on average, the extrema do not coincide exactly with the annotated ED and ES frames. Correcting for this systematic bias led to improved performance as shown in Table~\ref{tab:main}. However, because this correction requires a small subset of labelled data, the resulting method can no longer be considered fully unsupervised. 
The larger ED bias likely arises as ED occurs at the end of relaxation, when motion is minimal and adjacent frames differ only subtly. ES follows contraction, where the transition is sharper and fewer frames resemble the ES frame.
Since our phase signal is calculated from the zero-mean latent, $\hat{h}_t$, ED and ES appear maximally and conversely distinct from the mean, and hence occupy opposite ends of the orbit. 

\noindent\textbf{Conclusion.}\label{sec:conclusion}
We introduced a self-supervised method for cardiac phase detection based on the physiological prior that the cardiac cycle is governed by a single mechanical degree of freedom. By constraining the latent phase to a one-dimensional parametrised orbit the model predicts ED/ES frames unambiguously and without supervision after a fraction of the training time. Experiments on EchoNet-Dynamic show that this minimal representation outperforms baselines for ED/ES detection with strong cycle consistency, confirming that a principled inductive bias can outperform richer but less constrained representations. The continuous phase signal opens avenues for arrhythmia detection, cycle-resolved functional analysis, temporal synchronisation across patients, and zero-shot phase transfer to new views and modalities, with EF regression from the learned representations a first example of such downstream extensions.

\begin{credits}
\subsubsection{\ackname}

This work was supported by Ultromics Ltd., UK Research and Innovation [UKRI AI Centre for Doctoral Training in Digital Healthcare grant number EP/Y030974/1, UKRI Centre for Doctoral Training in AI for Healthcare grant number EP/S023283/1, and UKRI DTP award]. We acknowledge resources provided by AIRR, operated by the University of Bristol and funded by DSIT via UKRI and STFC [ST/AIRR/I-A-I/1023]~\cite{mcintoshsmith2024isambardai}. We used coding agents and LLMs from Anthropic and OpenAI for text and code polishing.

\subsubsection{\discintname}
The authors have no relevant competing interests.
\end{credits}

\bibliographystyle{splncs04}
\bibliography{references}

\end{document}